\pdfoutput=1

\documentclass[11pt]{article}

\usepackage[preprint]{acl}
\usepackage{xcolor} 
\usepackage{adjustbox}
\usepackage{times}
\usepackage{latexsym}
\usepackage{booktabs}
\usepackage{multirow}
\usepackage[T1]{fontenc}
\usepackage[utf8]{inputenc}
\usepackage{microtype}
\usepackage{inconsolata}
\usepackage{graphicx}
\usepackage{comment}
\usepackage{enumitem}
\usepackage{tabularx}
\usepackage{seqsplit}

\title{Chronological Passage Assembling in RAG framework for Temporal Question Answering}

\author{
Byeongjeong Kim,~Jeonghyun Park,~Joonho Yang,~Hwanhee Lee\thanks{Corresponding Author.} \\
    Department of Artificial Intelligence, Chung-Ang University\\
    \texttt{\{michael97k, tom0365, plm3332, hwanheelee\}@cau.ac.kr}
}

\begin{document}
\maketitle

\begin{abstract}
Long-context question answering over narrative tasks is challenging because correct answers often hinge on reconstructing a coherent timeline of events while preserving contextual flow in a limited context window. 
Retrieval-augmented generation (RAG) methods aim to address this challenge by selectively retrieving only necessary document segments. 
However, narrative texts possess unique characteristics that limit the effectiveness of these existing approaches. 
Specifically, understanding narrative texts requires more than isolated segments, as the broader context and sequential relationships between segments are crucial for comprehension. 
To address these limitations, we propose ChronoRAG, a novel RAG framework specialized for narrative texts. 
This approach focuses on two essential aspects: refining dispersed document information into coherent and structured passages and preserving narrative flow by explicitly capturing and maintaining the temporal order among retrieved passages. 
We empirically demonstrate the effectiveness of ChronoRAG through experiments on the NarrativeQA and GutenQA dataset, showing substantial improvements in tasks requiring both factual identification and comprehension of complex sequential relationships, underscoring that reasoning over temporal order is crucial in resolving narrative QA. \footnote{The source code will be released upon paper acceptance.}
\end{abstract}

\section{Introduction}
\label{lab:intro}
Long-context question answering tasks, which require the ability to utilize one or more long documents~\cite{pang2022quality}, present a significant challenge in natural language processing. While modern transformer-based Large Language Models (LLMs) have shown a remarkable ability to handle long contexts~\cite{liu2025comprehensive, wang2024ada}, they face fundamental limitations when confronted with extremely long-form text. Processing extensive documents for every query leads to major computational inefficiency, and as the context grows longer, the models' ability to accurately identify and prioritize relevant information decreases, impacting the reliability of their outputs.

To address these challenges, Retrieval-Augmented Generation (RAG)~\cite{lewis2020retrieval} has become a standard approach, focusing on efficiently retrieving only relevant segments from large documents to integrate into the model's context window. This selective retrieval method helps models leverage vast knowledge bases far beyond their built-in context limits.

\begin{figure}[!t]
\centering
\includegraphics[width=\linewidth]{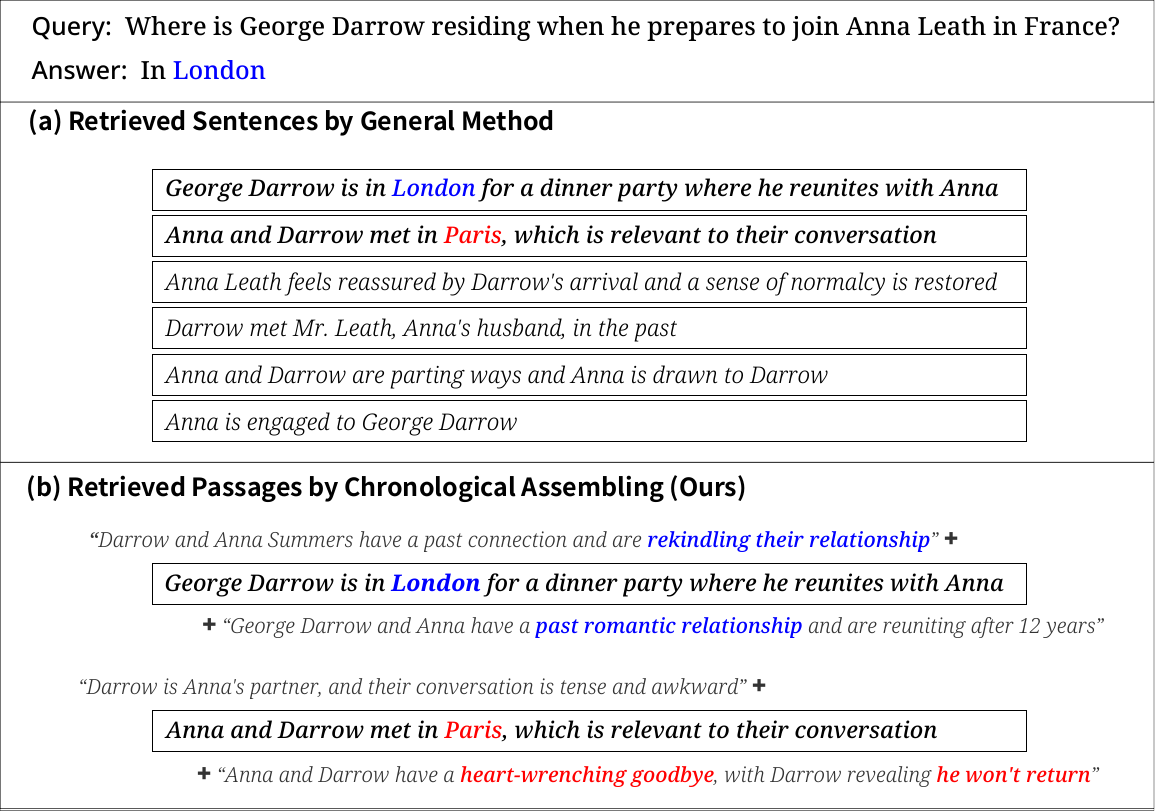}



\caption{Retrieval comparison for a narrative query. (a) Fine-grained indexing returns six standalone sentences, leaving key clues detached. (b) Our chronological assembling retrieves passages that include their immediate chronological context, preserving the narrative flow. Boxes indicate the directly retrieved sentences.}

\label{fig:intro}
\vspace{-4mm}
\end{figure}
However, a fundamental methodological gap exists in most RAG frameworks~\cite{lewis2020retrieval, sarthi2024raptor}: they primarily treat documents as a collection of short, independently-retrieved snippets of information. This methodology fundamentally conflicts with the sequential nature of long-form narratives, such as those found in history, literature, and film.
Narrative texts are uniquely defined by their structure; they can be \textbf{extremely long}, their individual passages often fail to convey the full story unless \textbf{read in order}, and grasping the \textbf{chronological and relational connections between passages is essential} for comprehension. Treating passages as isolated facts severs these critical links, fragmenting the narrative timeline.

Figure~\ref{fig:intro} illustrates the mismatch between conventional retrieval strategies and the characteristics of narrative data. As shown in (a) of Figure~\ref{fig:intro}, a common approach is to retrieve as many sentences as possible that are likely to match the query based on textual similarity. To do so, documents are typically stored as isolated sentences. While such methods may successfully retrieve a sentence containing the correct answer, they often fail to provide sufficient contextual cues. This can create ambiguity, making it unclear whether "London" or "Paris" is the location relevant to the question, even if both are mentioned in the retrieved results.

To address this issue, we introduce ChronoRAG, a novel RAG-based approach that embodies an alternative strategy grounded in the principle that solving narrative-based problems fundamentally requires recognizing the chronological order of events. Instead of maximizing the number of retrieved sentences, our framework, as shown in (b) of Figure~\ref{fig:intro}, retrieves fewer distinct informational units but includes their \textbf{surrounding context} to disambiguate meaning. This approach provides the crucial contextual clues—indicating that "London" is associated with a reunion while "Paris" pertains to a farewell—that are essential for accurate question answering. ChronoRAG achieves this by clarifying dispersed narrative content into structured passages and explicitly capturing the temporal relationships between them, enabling the retrieval of a coherent narrative flow rather than a collection of isolated facts.

We empirically validate our proposed approach on the NarrativeQA~\cite{kovcisky2018narrativeqa} and GutenQA~\cite{duarte-etal-2024-lumberchunker}. To rigorously test temporal reasoning, we isolate a subset of "Time Questions" that require understanding event sequences. Our experiments show that our method achieves significant improvements in both the complete dataset and the specialized Time Question set. Notably, these results are achieved using lighter graph construction and retrieval mechanisms than those found in existing summary and graph-based methods, demonstrating enhanced performance in identifying individual facts and comprehending complex relational structures.

Our contributions can be summarized as follows:
\begin{itemize}[leftmargin=*,itemsep=0.1em]

  \item We find that resolving narrative QA requires leveraging event chronology and preserving contextual flow, which guides our method in distilling dispersed story elements into coherent, temporally aware passages.
  \item We introduce a novel RAG framework, ChronoRAG, which refines raw text into structured passages, explicitly maintains temporal links between events, and incorporates adjacent context.
  \item Experiments demonstrate the effectiveness of our framework, and emphasizing event-to-event relations drives performance gains for both factual and temporal queries, highlighting the critical role of relational understanding over entity extraction.
\end{itemize}

\section{Related Work}

\textbf{Passage Granularity}
Document indexing approaches have been explored with varying passage granularity to improve retrieval precision. DenseXRetrieval~\cite{chen2024dense} advocates finer granularities to enhance information precision. Conversely, MolecularFacts~\cite{gunjal2024molecular} demonstrates that overly granular decompositions such as atomic facts or propositions often lose critical contextual cues, advocating instead for concise yet contextually coherent units. 
Our method strikes a balance by using atomic facts as keys for indexing, while preserving broader narrative flows as values during retrieval, thereby combining precision with coherence.

\paragraph{Summary-Based Document Augmentation}
Summary-based indexing methods construct hierarchical structures by iteratively compressing document segments into progressively shorter summaries. RAPTOR~\cite{sarthi2024raptor}, MemWalker~\cite{chen2023walking}, and ReadAgent~\cite{lee2024human} demonstrate how such designs enhance retrieval accuracy and contextual coherence by filtering out irrelevant passages and aggregating query-focused summaries. However, the deep hierarchies adopted in these approaches often lead to redundant overlaps and high computational costs. Our approach simplifies the hierarchical concept by adopting a single-layer summary, significantly reducing computation and overlap issues while maintaining contextual effectiveness.

\begin{figure*}[!t]
\centering
\includegraphics[width=\linewidth]{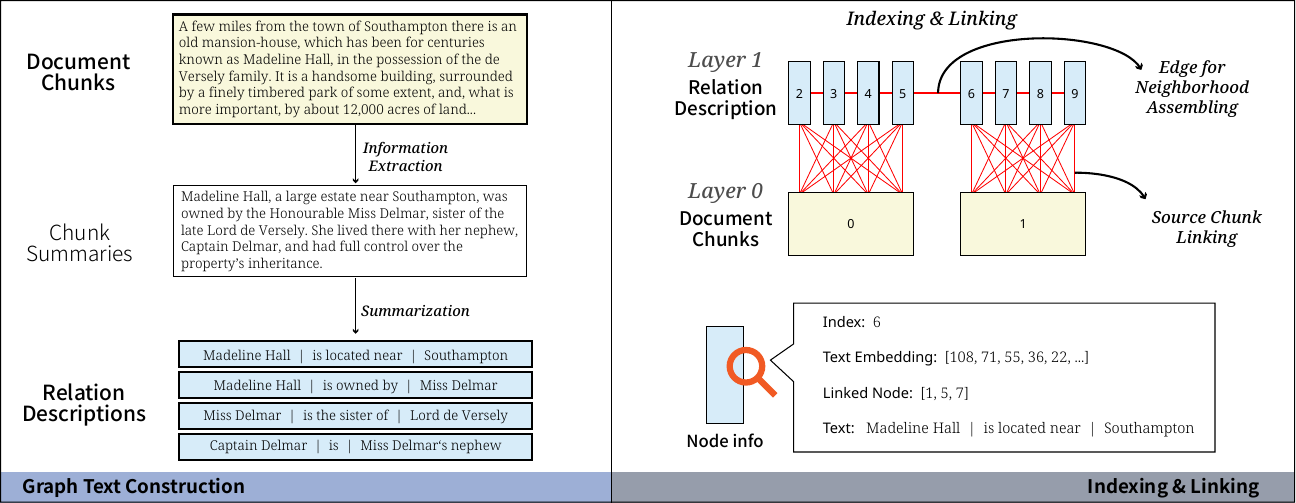}
\caption{The offline Graph Construction pipeline of ChronoRAG. This process transforms an unstructured narrative document into a structured, two-layer graph that explicitly encodes chronological relationships.}

\label{fig:method}
\vspace{-4mm}
\end{figure*}

\paragraph{Knowledge Graph-Based Document Augmentation}
Graph-based augmentation represents another line of work that emphasizes structured relational knowledge. GraphRAG~\cite{edge2024local} formalizes the paradigm through components such as query processors, retrievers, organizers, and generators, and leverages graph traversal and community detection to retrieve information beyond lexical similarity. LightRAG~\cite{guo2024lightrag} reduces preprocessing and latency overhead by encoding relational signals into dense indices and employing coarse-to-fine retrieval. Extensions such as EventRAG~\cite{yang2025eventrag} construct event knowledge graphs that capture temporal and causal dependencies, while Entity–Event RAG~\cite{zhang2025respecting} prevents collapsing distinct entity mentions by maintaining separate but linked entity and event subgraphs. Iterative reasoning approaches, such as KG-IRAG~\cite{yang2025beyond}, further refine this idea by incrementally retrieving over temporal and logical dependencies. Despite these advances, most methods emphasize static entity-centric relations; our framework differs by explicitly modeling sequential narrative relations, thereby addressing a critical gap in capturing dynamic contextual flows.

\section{ChronoRAG} 
\begin{figure*}[!t]
\centering
\includegraphics[width=\linewidth]{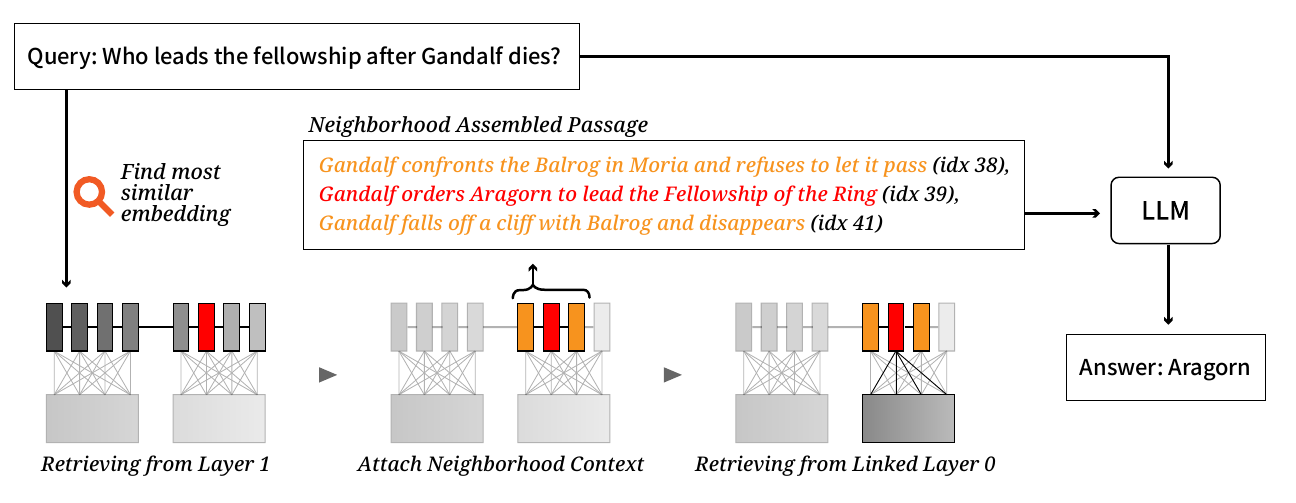}
\vspace{-2mm}
\caption{The online Passage Retrieval process of ChronoRAG for a sample query. This demonstrates how the pre-constructed graph is used at inference time to assemble a chronologically coherent context for the LLM.}
\label{fig:method_ret}
\vspace{-4mm}
\end{figure*}

We present ChronoRAG, a novel RAG framework specialized for narrative texts where chronological context is crucial. Most RAG systems treat documents as a collection of independent facts, which fragments the timeline and severs the contextual links essential for understanding sequential events. To address this, we design our framework to reconstruct narrative flow by explicitly modeling and preserving the temporal order of events. As described in Figure~\ref{fig:method}, our framework is composed of two primary stages: an offline \textbf{Graph Construction} phase where the original documents are processed into a hierarchical, linked structure, and an online \textbf{Passage Retrieval and Answer Generation} phase where the constructed graph is used to answer queries.

\label{sec:method}

\subsection{Offline Graph Construction}
This offline phase transforms a raw document into a structured, two-layer graph that captures both factual information and narrative chronology. As shown in Figure~\ref{fig:method}, this graph is composed of two levels: a foundational \textit{\textbf{Layer 0}} containing the original document text divided into sequential, fixed-length chunks, which preserves narrative detail, and an abstract \textit{\textbf{Layer 1}} built from concise, structured relation descriptions that represent the key events and relationships. The graph construction process involves four steps.

\subsubsection{Document Chunking}
Due to inherent limitations in processing an entire document simultaneously, we first divide the original document (\( D \)) into fixed-length chunks (\( d_i \)), each consisting of up to \( k \) tokens. This approach ensures that all retrieved document segments fit within a predefined context length, thereby maintaining both the quality and manageability of retrieval results. The document is initially segmented into individual sentences, which are then sequentially appended to each chunk. When the cumulative length exceeds \( k \) tokens, the next sentence is assigned to a new chunk, ensuring that sentences are not split across chunks. In rare cases where a single sentence itself exceeds \( k \) tokens, the sentence is split to guarantee that every chunk remains within the token limit. These segmented document chunks serve as the fundamental retrieval units and constitute \textit{Layer 0} of the graph constructed in subsequent stages. The document $D$ is thus represented as a set of chunks, where each chunk $d_i$ has a token count less than $k$, as follows:

\[
D = \{ d_1, d_2, d_3, \ldots , d_i\}, \quad |d_i| < k
\]

\subsubsection{Chunk Summarization}
Next, we sequentially cluster the chunks in the original document order, grouping every \( l \) chunks. We concatenate and summarize each cluster’s texts using an LLM. 
We utilize sequential clustering because it preserves the document’s original order while maintaining a controllable and consistent input length for the LLM. 
This summarization step facilitates higher-level representation learning by focusing on the overall flow of the document rather than retaining excessive local detail.
For example, the left panel of Figure~\ref{fig:method} shows how a raw \textit{Document Chunk} about \textit{Madeline Hall} is condensed into a more concise \textit{Chunk Summary}.
For each cluster of $l$ chunks, we generate a summary $S_i$ by an LLM using a summarization prompt $P_{\mathrm{summarize}}$ as formulated below: (We provide a full prompt in Appendix~\ref{app:prompt}.)

\[
s_i = \mathrm{LLM}(P_{\mathrm{summarize}}, \{ d_{i1}, d_{i2}, d_{i3}, \ldots, d_{il} \})
\]

\subsubsection{Entity-Relation Extraction}
We then transform the generated summaries via LLM into relational descriptions among entities. 
We adapt the prompt of GraphRAG~\cite{edge2024local} into a one-shot instruction for entity–relation extraction. (Full prompts are in Table~\ref{tab:prompt_main} of Appendix.)
From LLM's outputs, which consist of both entity descriptions($E_i$) and relation descriptions($R_i$), we only use the relation descriptions to form the \textit{Layer 1} nodes of the graph. 
This extraction step decomposes the summarized text into retrieval-friendly fragments, as described in the left panel of Figure~\ref{fig:method}, where the summary is broken down into atomic facts like \texttt{"Madeline Hall is owned by Miss Delmar"}.

Here, we exclude entity descriptions because identical entities may appear redundantly across multiple chunks. 
Finally, we formalize this extraction step, where an LLM processes each summary $S_i$ to produce a set of entities $E_i$ and relations $R_i$, as shown below: (Full prompts are in Appendix~\ref{app:prompt}.)

\[
\{ E_i, R_i \} = \mathrm{LLM}(P_{\mathrm{extraction}}, S_i)
\]

\subsubsection{Hierarchical and Temporal Indexing}
This step involves assigning indices to the relation description sentences ($R_i$) derived from the summary and the document chunks ($d_i$) from the original text. Document chunks are indexed sequentially according to their original order in the source text, while the relation description sentences are indexed either based on the earlier chunks from which they are derived and in the order in which they were generated during information extraction.

Each document chunk corresponds to a \textit{Layer-0} node, and each relation sentence forms a \textit{Layer-1} node.
For quick access, each \textit{Layer-1} node connects to its corresponding \textit{Layer-0} nodes—those within the cluster from which it was derived—by establishing directed edges. Additionally, adjacent \textit{Layer-1} nodes (according to their index) are also linked via edges.
This indexing and edge-construction process results in the formation of a unified graph structure.

\subsection{Online Passage Retrieval}
At inference time, we handle a query through a hierarchical retrieval process that leverages the constructed graph to assemble a rich, chronologically-aware context for the LLM.

\subsubsection{Hierarchical Retrieving}
We leverage the hierarchical granularity of \textit{Layer 1} and \textit{Layer 0} for retrieval. We begin by retrieving high-precision relation descriptions from \textit{Layer 1} based on semantic similarity to the query. Then, using the links established during indexing, we retrieve the related \textit{Layer 0} chunks to provide a comprehensive and balanced context. As illustrated in Figure~\ref{fig:method_ret}, this process first identifies a key event in \textit{Layer 1} and later retrieves the detailed source text from \textit{Layer 0}. This step is crucial because \textit{Layer 0} often retains omitted details and original dialogues that are valuable for question answering.

\subsubsection{Neighborhood Assembling}
We then augment retrieved relational descriptions with their surrounding context to reconstruct a narrative flow. Rather than relying on isolated facts, we aim to provide contextually rich information. 
As in the example of Figure~\ref{fig:method_ret}, after ChronoRAG retrieves a key event, such as \texttt{"Gandalf orders Aragorn to lead the Fellowship of the Ring (idx 39)"}, the system automatically appends its chronological neighbors, including \texttt{"Gandalf confronts the Balrog... (idx 38)"} and \texttt{"Gandalf falls off a cliff... (idx 41)"}. This creates a coherent, temporally ordered passage that preserves the local storyline, providing crucial context that isolated facts would lack.

\subsubsection{Answer Generation}
Finally, we combine the original query with the context obtained through hierarchical retrieval and neighborhood assembling and feed them into the language model. We separate each passage by double line breaks and sort by relevance, enabling accurate and coherent answer generation.
\section{Experiments}
\subsection{Experimental Setup}

\paragraph{Dataset} We employ the NarrativeQA~\cite{kovcisky2018narrativeqa} and GutenQA~\cite{duarte-etal-2024-lumberchunker} datasets to measure ChronoRAG's performance. NarrativeQA comprises 10,557 question-answer pairs from 391 stories, while GutenQA consists of 3,000 pairs from 1,000 stories. To specifically evaluate temporal reasoning, we construct a "Time Questions" subset by selecting all samples containing at least one of a set of predefined temporal keywords: \{`When,' `While,' `During,' `After,' `Before'\}. This process yields 1,111 questions from NarrativeQA and 662 from GutenQA, respectively. These questions require retrieving and reasoning over multiple related events, making them a demanding benchmark for temporal understanding.

\begin{table*}[ht]
\centering
\resizebox{\linewidth}{!}{
\begin{tabular}{l|cc|cc|cc}
\hline
\multicolumn{7}{c}{\textbf{NarrativeQA}} \\
\hline
Metric & \multicolumn{2}{c|}{ROUGE} &  \multicolumn{2}{c|}{CosineSim} & \multicolumn{2}{c}{LLM Eval (ACC)} \\
Subset & Whole Data & Time Question  & Whole Data & Time Question & Whole Data & Time Question \\ 
\hline
NaiveRAG   & 0.255   & 0.227   & 0.841 & 0.844  & 0.183 & 0.144 \\
Propositionizer & 0.262   & 0.238  & 0.846 & 0.852  & 0.189 & 0.141 \\
RAPTOR\_CT & \underline{0.298} & \underline{0.262} & \textbf{0.854} & \textbf{0.858} & \underline{0.241} & \underline{0.178} \\
RAPTOR\_TT & 0.289   & 0.253  & 0.851 & \underline{0.854}  & 0.231 & 0.170 \\
LightRAG   & 0.240   & 0.214 & 0.841 & 0.845  & 0.182 & 0.123 \\
GraphRAG   & 0.195   & 0.185  & 0.823 & 0.830  & 0.139 & 0.106 \\
ChronoRAG (Ours) & \textbf{0.308}  &  \textbf{0.268}  & \underline{0.853} & \underline{0.854} & \textbf{0.257} & \textbf{0.195} \\
\hline
\end{tabular}
}

\vspace{2mm}
\resizebox{\linewidth}{!}{
\begin{tabular}{l|cc|cc|cc}
\hline
\multicolumn{7}{c}{\textbf{GutenQA}} \\
\hline
Metric & \multicolumn{2}{c|}{ROUGE} &  \multicolumn{2}{c|}{CosineSim} & \multicolumn{2}{c}{LLM Eval (ACC)} \\
Subset & Whole Data & Time Question  & Whole Data & Time Question & Whole Data & Time Question \\ 
\hline
NaiveRAG   & \textbf{0.166} & \textbf{0.172}  & 0.769  & 0.778  &  \textbf{0.251} & \textbf{0.278} \\
Propositionizer & 0.151 & 0.142  & \textbf{0.779}  & \textbf{0.785}  & 0.167 & 0.140 \\
RAPTOR\_CT & \underline{0.159} & 0.164  &  0.775 &  \underline{0.784} & 0.244 & 0.269 \\
RAPTOR\_TT & 0.104 & 0.102  &  0.762 &  0.769 & 0.134 & 0.119 \\
LightRAG   & 0.083 & 0.083 & 0.740  & 0.750  & 0.075 & 0.077 \\
GraphRAG   & 0.122 & 0.115  &  0.760 &  0.767 & 0.134 & 0.119 \\
ChronoRAG (Ours) & \underline{0.159} & \underline{0.170}  &   \underline{0.776} & \textbf{0.785}  &  \underline{0.248} & \underline{0.275}  \\
\hline
\end{tabular}
}
\caption{QA Performance on NarrativeQA and GutenQA across ROUGE, CosineSim, and LLM Eval metrics. Top performance is bolded, Second best is underlined.}
\label{tab:main-results}
\vspace{-3mm}
\end{table*}

\paragraph{Evaluation Metric} We measure answer quality using ROUGE-L~\cite{lin2004rouge}, which computes the Longest Common Subsequence (LCS) overlap between a generated answer and its corresponding human reference. Due to the short and pronoun-heavy nature of NarrativeQA answers, ROUGE-L effectively captures agreement in key word sequences without penalizing minor rephrasings.

In addition to ROUGE-L, we employ cosine similarity and LLM-based evaluation to better assess semantic fidelity. ROUGE may fail to capture semantically similar answers that differ lexically, so we incorporate complementary metrics to address this limitation. We compute cosine similarity using the Snowflake model~\cite{merrick2024arctic}, which is also used during the retrieval process. It measures the embedding-based similarity between the generated answer and the reference answer.

For LLM-based evaluation, we use GPT-4.1 Mini~\cite{achiam2023gpt}. Given the question, summary, gold passage, and ground-truth answer, the GPT model performs binary classification—[Correct] or [Wrong]—to determine whether the generated answer can be considered valid. See Appendix~\ref{app:prompt} for the detailed prompt.

\paragraph{Baselines} We compare against five existing methods that differ in information extraction, representation, and retrieval structure:
\begin{itemize}[leftmargin=*,itemsep=0.1em]
  \item \textbf{NaiveRAG:} A standard RAG pipeline that performs chunk-level retrieval only, without further structuring~\cite{lewis2020retrieval}.
  \item \textbf{RAPTOR:} Clusters semantically similar chunks via embedding similarity and builds a recursive summarization tree over clusters to guide retrieval—CT (Collapsed Tree) flattens each root-to-leaf path into one high-level summary, whereas TT (Tree Traversal) retains the full hierarchy and drills down level-by-level to gather finer-grained context~\cite{sarthi2024raptor}.
  \item \textbf{LightRAG:} Constructs a lightweight entity–relation graph to enable fast context retrieval using dual-level extraction, prioritizing computational efficiency and incremental updates~\cite{guo2024lightrag}.
  \item \textbf{GraphRAG:} Builds a richer graph with detailed relation weighting and neighborhood assembly to support deeper multi-hop retrieval, capturing both high-level relation summaries and their underlying chunks~\cite{edge2024local}.
  \item \textbf{Propositionizer:} Transforms the entire source text into fine-grained propositions (atomic sentences) and treats each proposition as a retrieval unit, then feeds retrieved propositions into the generation model~\cite{chen2024dense}.
\end{itemize}

\paragraph{Implementation Details.} All baselines share identical hyperparameter settings: top-k of 20 for retrieval, contextTokenLengthLimit of 1,500 tokens, and the same greedy decoding strategy during generation. We perform all summarization and entity–relation extraction steps with meta-llama-3-8B-Instruct~\cite{grattafiori2024llama}. We compute retrieval scores using embedding similarity exclusively; we don't use BM25~\cite{robertson2009probabilistic} to prevent distortion of the original text during generation. Specifically, we employ the arctic-Snowflake-embed-l~\cite{merrick2024arctic} for generating embeddings, and use unifiedqa-v2-t5-3b-1363200~\cite{khashabi2022unifiedqa} for final answer generation. All retrieved contexts fed into the generator respect the 1,500-token length limit to ensure fair comparison across all methods.


\subsection{Main Results}


\paragraph{Performance Comparison}
Table~\ref{tab:main-results} shows that our proposed ChronoRAG outperforms all baselines on the NarrativeQA dataset, with particularly strong gains on the "Time Question" subset. An analysis of the baselines on this dataset reveals distinct failure modes corresponding to their retrieval strategies. Summarization-based methods like RAPTOR-CT are the next-best performers but still lag our method; RAPTOR's approach of clustering semantically similar chunks is insufficient for narratives, as it can group thematically related but chronologically distant events. Similarly, methods retrieving isolated text units, such as NaiveRAG and Propositionizer, struggle to provide sufficient context and fragment the narrative flow. GraphRAG records the lowest score, as its exhaustive entity–relation extraction adds thousands of trivial nodes, burying key plot elements under noise and severely diluting precision.

However, on the GutenQA dataset, the results are more nuanced. We explain that this is likely due to the nature of questions in GutenQA, which are often constructed by extracting text directly from the source and thus reward methods with the most direct access to the original passage details. For this reason, methods that heavily refine or abstract the text, such as Propositionizer and the graph-based approaches, perform poorly because they lose the specific passage-level information required. Even RAPTOR is penalized, as its summary-first approach limits direct access to the source text. In contrast, both NaiveRAG and ChronoRAG employ a consistent, fixed-length chunking strategy, which helps normalize the potentially unrefined structure of the GutenQA data. NaiveRAG’s slight edge in some metrics can be attributed to its direct retrieval from these unaltered chunks, whereas ChronoRAG's abstraction steps, while beneficial for narrative synthesis, risk filtering out the fine-grained details that these specific questions demand.

\begin{figure*}[!ht]
\centering
\includegraphics[width=0.85\linewidth]{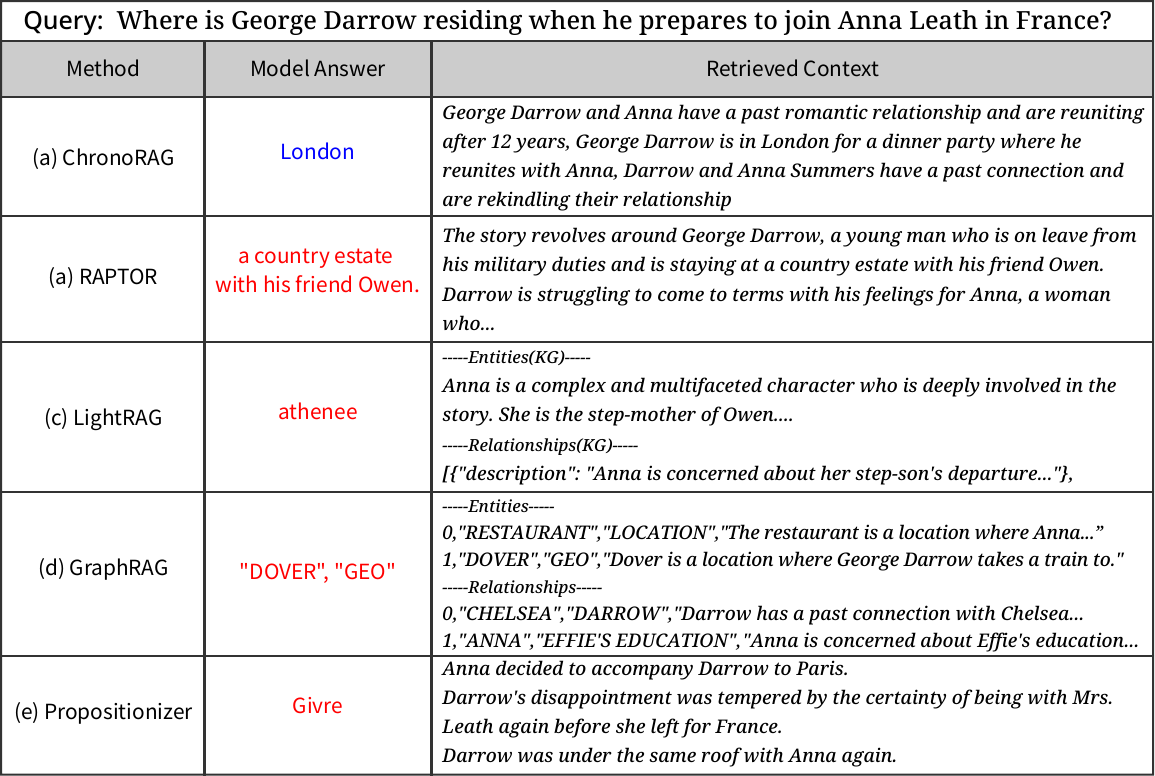}
\caption{A qualitative comparison of retrieved context and model answers for the query, "Where is George Darrow residing when he prepares to join Anna Leath in France?".}
\label{fig:case}
\vspace{-4mm}
\end{figure*}

\paragraph{Ablation Study}
We conduct ablation studies to investigate the effectiveness of different components and settings of ChronoRAG. As shown in the Table~\ref{tab:ablation}, without summarizing the original text and extracting entity relations shows a significant performance degradation, showing the importance of chunk summarization. The effects of summarization are twofold: it leaves only important information, making retrieval easier, and when assembling, it clarifies the flow.
The results without passage assembling are obtained by individually searching for entity relations extracted from the summary, while the results without chunk summarization are obtained by searching for entity relations directly extracted from the 10 chunks. Despite not connecting nearby passages in both settings, a significant performance difference is observed in the TimeQuestion.

\begin{table}[ht]
\centering
\small
\renewcommand{\arraystretch}{1.5} 
\begin{tabular*}{\columnwidth}{l @{\extracolsep{\fill}} cc}
\hline
\textbf{Method} & \textbf{Whole Data} & \textbf{Time Question} \\ \hline
ChronoRAG                 & \textbf{0.308}      & \textbf{0.268}         \\
w/o Passage Assembling    & 0.295               & 0.252                  \\
w/o Chunk Summarization   & 0.272               & 0.233                  \\
w/o Relation Extraction   & 0.255               & 0.227                  \\ \hline
\end{tabular*}
\caption{Ablation study of ChronoRAG's core components on the NarrativeQA dataset, with performance measured by ROUGE-L.}
\label{tab:ablation}
\vspace{-3mm}
\end{table}

\subsection{Analysis}

\paragraph{Trade-off between Linking Window and the Number of Retrieved Passage}
We analyze the trade-off of using a larger linking window for passage assembly. While a wider window provides more local context, it also reduces the number of distinct passages that can be retrieved within a fixed token budget. Our experiment confirms this is detrimental; as shown in Table~\ref{tab:vary}, extending the window to two neighbors ("Extended Link Window") lowers performance on both the whole dataset and the temporal questions subset. This result validates that our default approach of using a more concise, immediately adjacent context is more effective.

\begin{table}[ht]
\centering
\small
\renewcommand{\arraystretch}{1.5} 
\begin{tabular*}{\columnwidth}{l @{\extracolsep{\fill}} cc}
\hline
\textbf{NarrativeQA}      & \textbf{Whole Data} & \textbf{Time Question} \\ \hline
ChronoRAG                 & \textbf{0.308}      & \textbf{0.268}         \\
Extended Link Window      & 0.300               & 0.258                  \\
Merged Key                & 0.302               & 0.257                  \\ \hline
\end{tabular*}
\caption{Analysis of design variations within the ChronoRAG framework on the NarrativeQA dataset measured by ROUGE-L.}
\vspace{-5mm}
\label{tab:vary}
\end{table}

\paragraph{Key-Value Separation in Information Retrieval}
We investigate the effectiveness of key-value retrieval design of ChronoRAG, which separates the precise fact used for retrieval (the key) from the broader context provided to the model (the value). To validate this, we test an alternative "Merged Key" approach where the retrieved fact and its neighbors are combined into a single text unit before retrieval. As shown in Table~\ref{tab:vary}, this modification results in a slight performance decrease, indicating that our key-value separation is an effective strategy for balancing retrieval precision and contextual coherence.

\paragraph{Effect of Neighborhood Assembling}

\begin{table}[ht]
\centering
\small
\renewcommand{\arraystretch}{1.5} 
\begin{tabular*}{\columnwidth}{l @{\extracolsep{\fill}} cc}
\hline      & \textbf{\# of Sentence} & \textbf{mean(Similarity)    } \\ \hline
Retrieved Sentences                 &  104{,}981      & 0.838        \\
Assembled Sentences      & 209{,}092               & 0.785                  \\\hline
\end{tabular*}
\caption{Embedding Similarity Between Query and Retrieved/Assembled Passages.}
\label{tab:assem_sim}
\vspace{-3mm}
\end{table}

We validate the effectiveness of neighborhood assembling in enriching the retrieved context with information that is chronologically relevant but not necessarily the most semantically similar to the query. To demonstrate this, we compare the average embedding similarity between the query and the initially retrieved sentences versus the final assembled passages. As shown in Table~\ref{tab:assem_sim}, the average similarity for the assembled passages is discernibly lower than that of the sentences retrieved by similarity alone. This gap indicates that the neighboring passages, while chronologically adjacent, are semantically distinct from the initial query hit. In narrative texts where surrounding content carries strong causal or temporal relevance, this mechanism allows the model to incorporate pertinent information beyond the limits of pure similarity search, thereby improving the context for answer generation.

\paragraph{Case Study}
Figure~\ref{fig:case} presents excerpts of the original passages retrieved by each method for the example shown in Figure~\ref{fig:intro}. 
RAPTOR retrieves summary passages, which enable access to content covering a wide range of information. 
However, these summaries frequently include information that is not pertinent to the query, or conversely, omit critical details necessary for answering the question due to length constraints imposed by the summarization process.
LightRAG and GraphRAG extract entities and relations directly from the original text. In particular, GraphRAG was found to underperform compared to direct retrieval to the source chunk, likely due to its tendency to include exhaustive explanations of all elements. 
Propositionizer and LightRAG offer relatively general-level granularity explanations, yet they still struggle to address questions that require understanding the changes in the relationship between Anna and George.
In contrast, ChronoRAG identifies the minimal set of chronologically adjacent passages while suppressing unrelated narrative details, illustrating its strength in maintaining temporal coherence and reducing retrieval noise.

\paragraph{Computation Costs}
Our pipeline is computationally efficient, requiring just two LLM calls per 1,000 tokens for graph construction. Although this cost increases linearly with document length, it remains lower than competing methods like recursive summarization. Furthermore, only one LLM call is required for answer generation during search, with our method still attaining the highest performance despite its efficiency.

\section{Conclusion}

We present ChronoRAG, an RAG framework that can effectively and efficiently handle narrative text. Our framework refines content through summarization and relation extraction, and improves overall performance through simple passage augmentation that connects adjacent events via an index.
This suggests that it is important not only to organize individual events and elements in narrative texts but also to connect events that are spatially and temporally close to each other.
\section*{Limitations}
Our proposed ChronoRAG framework explicitly models temporal order to improve narrative question answering. However, the approach has several limitations. First, while our graph construction pipeline is lightweight compared to prior graph-based methods, it still requires multiple LLM calls for summarization and relation extraction, which may introduce latency in large-scale deployments. Second, our evaluation focuses primarily on two English narrative datasets (NarrativeQA and GutenQA), and results may not directly generalize to non-English narratives or other domains such as legal or medical texts. Third, although our method improves temporal reasoning, it does not yet capture more complex discourse phenomena such as causal chains spanning distant events. Future work could explore integrating richer discourse structures and extending experiments to multilingual or domain-specific corpora. 

\section*{Ethics Statement}
All experiments in this paper are conducted on publicly available datasets (NarrativeQA and GutenQA) and widely used pre-trained models under their respective licenses. No private or sensitive user data is involved. While narrative corpora are relatively low-risk, they may still reflect historical or cultural biases present in the source texts, which can propagate into retrieval and generation. We report results in aggregate without attempting to infer personal attributes, and we adhere to ethical guidelines for reproducible research by ensuring transparency in data usage and methodology. We also plan to release our implementation to support open and verifiable research practices. 



\section*{Acknowledgments}
This research was supported by Institute for Information \& Communications Technology Planning \& Evaluation (IITP) through the Korea government (MSIT) under Grant No. 2021-0-01341 (Artificial Intelligence Graduate School Program (Chung-Ang University)).

\bibliography{acl_latex}

\begin{thebibliography}{22}
\providecommand{\natexlab}[1]{#1}

\bibitem[{Achiam et~al.(2023)Achiam, Adler, Agarwal, Ahmad, Akkaya, Aleman, Almeida, Altenschmidt, Altman, Anadkat et~al.}]{achiam2023gpt}
Josh Achiam, Steven Adler, Sandhini Agarwal, Lama Ahmad, Ilge Akkaya, Florencia~Leoni Aleman, Diogo Almeida, Janko Altenschmidt, Sam Altman, Shyamal Anadkat, et~al. 2023.
\newblock Gpt-4 technical report.
\newblock \emph{arXiv preprint arXiv:2303.08774}.

\bibitem[{Chen et~al.(2023)Chen, Pasunuru, Weston, and Celikyilmaz}]{chen2023walking}
Howard Chen, Ramakanth Pasunuru, Jason Weston, and Asli Celikyilmaz. 2023.
\newblock Walking down the memory maze: Beyond context limit through interactive reading.
\newblock \emph{arXiv preprint arXiv:2310.05029}.

\bibitem[{Chen et~al.(2024)Chen, Wang, Chen, Yu, Ma, Zhao, Zhang, and Yu}]{chen2024dense}
Tong Chen, Hongwei Wang, Sihao Chen, Wenhao Yu, Kaixin Ma, Xinran Zhao, Hongming Zhang, and Dong Yu. 2024.
\newblock Dense x retrieval: What retrieval granularity should we use?
\newblock In \emph{Proceedings of the 2024 Conference on Empirical Methods in Natural Language Processing}, pages 15159--15177.

\bibitem[{Duarte et~al.(2024)Duarte, Marques, Gra{\c{c}}a, Freire, Li, and Oliveira}]{duarte-etal-2024-lumberchunker}
Andr{\'e}~V. Duarte, Jo{\~a}o~DS Marques, Miguel Gra{\c{c}}a, Miguel Freire, Lei Li, and Arlindo~L. Oliveira. 2024.
\newblock \href {https://doi.org/10.18653/v1/2024.findings-emnlp.377} {{L}umber{C}hunker: Long-form narrative document segmentation}.
\newblock In \emph{Findings of the Association for Computational Linguistics: EMNLP 2024}, pages 6473--6486, Miami, Florida, USA. Association for Computational Linguistics.

\bibitem[{Edge et~al.(2024)Edge, Trinh, Cheng, Bradley, Chao, Mody, Truitt, Metropolitansky, Ness, and Larson}]{edge2024local}
Darren Edge, Ha~Trinh, Newman Cheng, Joshua Bradley, Alex Chao, Apurva Mody, Steven Truitt, Dasha Metropolitansky, Robert~Osazuwa Ness, and Jonathan Larson. 2024.
\newblock From local to global: A graph rag approach to query-focused summarization.
\newblock \emph{arXiv preprint arXiv:2404.16130}.

\bibitem[{Grattafiori et~al.(2024)Grattafiori, Dubey, Jauhri, Pandey, Kadian, Al-Dahle, Letman, Mathur, Schelten, Vaughan et~al.}]{grattafiori2024llama}
Aaron Grattafiori, Abhimanyu Dubey, Abhinav Jauhri, Abhinav Pandey, Abhishek Kadian, Ahmad Al-Dahle, Aiesha Letman, Akhil Mathur, Alan Schelten, Alex Vaughan, et~al. 2024.
\newblock The llama 3 herd of models.
\newblock \emph{arXiv preprint arXiv:2407.21783}.

\bibitem[{Gunjal and Durrett(2024)}]{gunjal2024molecular}
Anisha Gunjal and Greg Durrett. 2024.
\newblock Molecular facts: Desiderata for decontextualization in llm fact verification.
\newblock In \emph{Findings of the Association for Computational Linguistics: EMNLP 2024}, pages 3751--3768.

\bibitem[{Guo et~al.(2024)Guo, Xia, Yu, Ao, and Huang}]{guo2024lightrag}
Zirui Guo, Lianghao Xia, Yanhua Yu, Tu~Ao, and Chao Huang. 2024.
\newblock Lightrag: Simple and fast retrieval-augmented generation.
\newblock \emph{arXiv preprint arXiv:2410.05779}.

\bibitem[{Khashabi et~al.(2022)Khashabi, Kordi, and Hajishirzi}]{khashabi2022unifiedqa}
Daniel Khashabi, Yeganeh Kordi, and Hannaneh Hajishirzi. 2022.
\newblock Unifiedqa-v2: Stronger generalization via broader cross-format training.
\newblock \emph{arXiv preprint arXiv:2202.12359}.

\bibitem[{Ko{\v{c}}isk{\`y} et~al.(2018)Ko{\v{c}}isk{\`y}, Schwarz, Blunsom, Dyer, Hermann, Melis, and Grefenstette}]{kovcisky2018narrativeqa}
Tom{\'a}{\v{s}} Ko{\v{c}}isk{\`y}, Jonathan Schwarz, Phil Blunsom, Chris Dyer, Karl~Moritz Hermann, G{\'a}bor Melis, and Edward Grefenstette. 2018.
\newblock The narrativeqa reading comprehension challenge.
\newblock \emph{Transactions of the Association for Computational Linguistics}, 6:317.

\bibitem[{Lee et~al.(2024)Lee, Chen, Furuta, Canny, and Fischer}]{lee2024human}
Kuang-Huei Lee, Xinyun Chen, Hiroki Furuta, John Canny, and Ian Fischer. 2024.
\newblock A human-inspired reading agent with gist memory of very long contexts.
\newblock In \emph{Proceedings of the 41st International Conference on Machine Learning}, pages 26396--26415.

\bibitem[{Lewis et~al.(2020)Lewis, Perez, Piktus, Petroni, Karpukhin, Goyal, K{\"u}ttler, Lewis, Yih, Rockt{\"a}schel et~al.}]{lewis2020retrieval}
Patrick Lewis, Ethan Perez, Aleksandra Piktus, Fabio Petroni, Vladimir Karpukhin, Naman Goyal, Heinrich K{\"u}ttler, Mike Lewis, Wen-tau Yih, Tim Rockt{\"a}schel, et~al. 2020.
\newblock Retrieval-augmented generation for knowledge-intensive nlp tasks.
\newblock \emph{Advances in neural information processing systems}, 33:9459--9474.

\bibitem[{Lin(2004)}]{lin2004rouge}
Chin-Yew Lin. 2004.
\newblock Rouge: A package for automatic evaluation of summaries.
\newblock In \emph{Text summarization branches out}, pages 74--81.

\bibitem[{Liu et~al.(2025)Liu, Zhu, Bai, He, Liao, Que, Wang, Zhang, Zhang, Zhang et~al.}]{liu2025comprehensive}
Jiaheng Liu, Dawei Zhu, Zhiqi Bai, Yancheng He, Huanxuan Liao, Haoran Que, Zekun Wang, Chenchen Zhang, Ge~Zhang, Jiebin Zhang, et~al. 2025.
\newblock A comprehensive survey on long context language modeling.
\newblock \emph{arXiv preprint arXiv:2503.17407}.

\bibitem[{Merrick et~al.(2024)Merrick, Xu, Nuti, and Campos}]{merrick2024arctic}
Luke Merrick, Danmei Xu, Gaurav Nuti, and Daniel Campos. 2024.
\newblock Arctic-embed: Scalable, efficient, and accurate text embedding models.
\newblock \emph{arXiv preprint arXiv:2405.05374}.

\bibitem[{Pang et~al.(2022)Pang, Parrish, Joshi, Nangia, Phang, Chen, Padmakumar, Ma, Thompson, He et~al.}]{pang2022quality}
Richard~Yuanzhe Pang, Alicia Parrish, Nitish Joshi, Nikita Nangia, Jason Phang, Angelica Chen, Vishakh Padmakumar, Johnny Ma, Jana Thompson, He~He, et~al. 2022.
\newblock Quality: Question answering with long input texts, yes!
\newblock In \emph{2022 Conference of the North American Chapter of the Association for Computational Linguistics: Human Language Technologies, NAACL 2022}, pages 5336--5358. Association for Computational Linguistics (ACL).

\bibitem[{Robertson et~al.(2009)Robertson, Zaragoza et~al.}]{robertson2009probabilistic}
Stephen Robertson, Hugo Zaragoza, et~al. 2009.
\newblock The probabilistic relevance framework: Bm25 and beyond.
\newblock \emph{Foundations and Trends{\textregistered} in Information Retrieval}, 3(4):333--389.

\bibitem[{Sarthi et~al.(2024)Sarthi, Abdullah, Tuli, Khanna, Goldie, and Manning}]{sarthi2024raptor}
Parth Sarthi, Salman Abdullah, Aditi Tuli, Shubh Khanna, Anna Goldie, and Christopher~D. Manning. 2024.
\newblock Raptor: Recursive abstractive processing for tree-organized retrieval.
\newblock In \emph{International Conference on Learning Representations (ICLR)}.

\bibitem[{Wang et~al.(2024)Wang, Duan, Zhang, Lin, and Chen}]{wang2024ada}
Chonghua Wang, Haodong Duan, Songyang Zhang, Dahua Lin, and Kai Chen. 2024.
\newblock Ada-leval: Evaluating long-context llms with length-adaptable benchmarks.
\newblock In \emph{Proceedings of the 2024 Conference of the North American Chapter of the Association for Computational Linguistics: Human Language Technologies (Volume 1: Long Papers)}, pages 3712--3724.

\bibitem[{Yang et~al.(2025{\natexlab{a}})Yang, Xue, Razzak, Hacid, and Salim}]{yang2025beyond}
Ruiyi Yang, Hao Xue, Imran Razzak, Hakim Hacid, and Flora~D Salim. 2025{\natexlab{a}}.
\newblock Beyond single pass, looping through time: Kg-irag with iterative knowledge retrieval.
\newblock \emph{arXiv preprint arXiv:2503.14234}.

\bibitem[{Yang et~al.(2025{\natexlab{b}})Yang, Wang, Shi, Yao, Liang, Ding, Yilmaz, Chen, and Zhang}]{yang2025eventrag}
Zairun Yang, Yilin Wang, Zhengyan Shi, Yuan Yao, Lei Liang, Keyan Ding, Emine Yilmaz, Huajun Chen, and Qiang Zhang. 2025{\natexlab{b}}.
\newblock Eventrag: Enhancing llm generation with event knowledge graphs.
\newblock In \emph{Proceedings of the 63rd Annual Meeting of the Association for Computational Linguistics (Volume 1: Long Papers)}, pages 16967--16979.

\bibitem[{Zhang et~al.(2025)Zhang, Li, Li, Ding, and Low}]{zhang2025respecting}
Ze~Yu Zhang, Zitao Li, Yaliang Li, Bolin Ding, and Bryan Kian~Hsiang Low. 2025.
\newblock Respecting temporal-causal consistency: Entity-event knowledge graphs for retrieval-augmented generation.
\newblock \emph{arXiv preprint arXiv:2506.05939}.

\end{thebibliography}
\clearpage
\appendix

\section{The use of Large Language Models}
We prepared the manuscript independently and used an LLM assistant solely for minor refinement purposes (e.g., clarity improvements and grammar checking). The assistant was not involved in research ideation or content creation. The tool we employed was ChatGPT-5.

\section{Implementation Details}
We conduct our experiments using an AMD EPYC 7313 CPU (3.0 GHz) paired with four NVIDIA RTX 4090 GPUs. We use Python 3.11.5 and PyTorch 2.3.1 for the software environment.
We access meta-llama-3-8B-Instruct via the OpenRouter (2025) API with temperature set to 0 (greedy decoding) for generating answers from GutenQA. The detailed hyperparameters used in our experiments can be found in Table~\ref{tab:model_config}.

We utilize the nano-graphrag repository\footnote{\url{https://github.com/gusye1234/nano-graphrag}} and the lightrag repository\footnote{\url{https://github.com/HKUDS/LightRAG}}
 to implement the GraphRAG and LightRAG baselines, respectively. For proposition generation from the original documents, we employ the chentong00/propositionizer-wiki-flan-t5-large model.

In our experiments, GraphRAG is configured to operate in local mode, while LightRAG is set to hybrid mode. For both models, the retrieval parameter top\_k is set to 10.

\begin{table*}[ht]
\centering
\renewcommand{\arraystretch}{1.2} 
\begin{tabular}{l|l}
\toprule
\textbf{Parameter} & \textbf{Value} \\ 
\midrule
\textbf{max token length in chunk} & 100 \\ 
\textbf{number of cluster in chunk} & 10 \\ 
\textbf{max token length in summarization} & 2000 \\ 
\textbf{max token length in Entity Relation Extraction} & 2000 \\ 
\textbf{do\_sample} & False \\ 
\textbf{summarization model} & meta-llama/Llama-3.1-8B-Instruct \\ 
\textbf{entity relation extraction model} & meta-llama/Llama-3.1-8B-Instruct \\ 
\textbf{model for answer generation} & unifiedqa-v2-t5-3b- 4281363200 \\ 
\textbf{embedding model for text embedding} & Snowflake/snowflake-arctic-embed-l \\ 
\textbf{max token length of context} & 15 \\ 
\textbf{retrieving top\_k} & 15 \\ 
\textbf{max number of retrieved passage} & 20 \\ 
\bottomrule
\end{tabular}
\caption{Configuration parameters for ChronoRAG pipeline.}
\label{tab:model_config}
\end{table*}

\section{Prompts}
\label{app:prompt}
\paragraph{ChronoRAG} As shown in Table~\ref{tab:prompt_main}, we employ task-specific prompts for both summarization and entity--relation extraction. 
For summarization, the model is instructed to condense each cluster of document chunks into a concise description without pronouns, ensuring that the resulting text remains self-contained. 
For entity--relation extraction, the model is guided by a structured instruction that requires listing entities with type and description, followed by explicit relationships between entities with a numeric strength score. 
This structured output is essential for constructing the ChronoRAG graph, as it enables us to represent both the factual content of the story and the temporal or relational dependencies between entities. 
By combining these two prompts, we distill long narrative texts into coherent graph structures that support accurate and temporally consistent retrieval.

\begin{table*}[t]
\centering
\renewcommand{\arraystretch}{1.12}
\setlength{\tabcolsep}{4pt}
\begin{tabularx}{\linewidth}{>{\raggedright\arraybackslash}X}
\toprule
\textbf{Summarization Prompt} \\
\cmidrule(lr){1-1}
System: Write a summary of the following context as short as possible within five sentences. DO NOT USE PRONOUN. \\
Context: \texttt{<document chunk>} \\
Summary: \\[0.6em]

\cmidrule(lr){1-1}
\textbf{Entity--Relation Extraction Prompt} \\
\cmidrule(lr){1-1}

\textbf{Goal} \\
Given a text document that is potentially relevant to this activity and a list of entity types, identify all entities of those types from the text and all relationships among the identified entities. \\[0.4em]
\cmidrule(lr){1-1}

\textbf{Steps} \\
1.\;Identify all entities. For each identified entity, extract: \\
\hspace{1em}-- \texttt{entity\_name}: Name of the entity, capitalized \\
\hspace{1em}-- \texttt{entity\_type}: One of \texttt{[Leading Role, Supporting Role, Object]} \\
\hspace{1em}-- \texttt{entity\_description}: Comprehensive description of the entity's attributes and activities \\
\textit{Format:} \texttt{("entity"|<entity\_name>|<entity\_type>|<entity\_description>)} \\[0.35em]
2.\;From step~1 entities, identify clearly related (source, target) pairs and extract: \\
\hspace{1em}-- \texttt{source\_entity}, \texttt{target\_entity} \\
\hspace{1em}-- \texttt{relationship\_description} \\
\hspace{1em}-- \texttt{relationship\_strength} (numeric) \\
\textit{Format:} \texttt{("relationship"|<source\_entity>|<target\_entity>|<relationship\_description> |<relationship\_strength>)} \\[0.35em]
3.\;Return all items in English as a single list delimited by \texttt{\&}. \\
4.\;Finish with \texttt{<End>}. \\[0.6em]
\cmidrule(lr){1-1}

\textbf{Example} \\
\textit{Text:} The Verdantis's Central Institution will meet on Monday and Thursday; a policy decision is due Thursday 1{:}30\,p.m. PDT, followed by a press conference where Chair Martin Smith will take questions. Investors expect the Market Strategy Committee to hold the benchmark rate at 3.5\%--3.75\%. \\[0.35em]
\textit{Output:} \\
\begin{minipage}{\linewidth}\ttfamily
("entity"|CENTRAL INSTITUTION|ETC|The Central Institution is the Federal Reserve of Verdantis, setting interest rates on Monday and Thursday)\\
\& ("entity"|MARTIN SMITH|PERSON|Chair of the Central Institution)\\
\& ("entity"|MARKET STRATEGY COMMITTEE|ORGANIZATION|Committee making key decisions about interest rates)\\
\& ("relationship"|MARTIN SMITH|CENTRAL INSTITUTION|Chair will answer questions at a press conference|9)\\
<End>
\end{minipage} \\[0.6em]
\cmidrule(lr){1-1}

\textbf{Real Data} \\
\textit{Text:} \texttt{<summary text>} \hfill \textit{Output:} \\
\bottomrule
\end{tabularx}
\caption{Prompt templates used in ChronoRAG for summarization and entity--relation extraction.}
\label{tab:prompt_main}
\end{table*}

\paragraph{LLM Eval} In addition, as shown in Table~\ref{tab:prompt_narr} and Table~\ref{tab:prompt_guten}, we assess model outputs with an LLM-based judge. GPT-4.1 Mini is prompted with a structured template that includes the question, a gold passage providing narrative context, the gold answer, and the model-generated answer. The prompt explicitly instructs the judge to output only one of two labels—\texttt{[Correct]} or \texttt{[Wrong]}—without explanation. This design ensures consistency and avoids subjective variation. The evaluation complements automatic metrics such as ROUGE and embedding-based similarity by correctly recognizing semantically valid answers even when they differ lexically.
\begin{table*}[t]
\centering
\begin{tabular}{p{\linewidth}}
\toprule
\textbf{System Prompt} \\
\midrule
You are the grader who determines the correct answer precisely. \\
\midrule
\textbf{User Prompt} \\
\midrule
Determine whether the user's answer to the following question is correct or not.\\
Choose only one of the two options: \texttt{[Correct]} or \texttt{[Wrong]}.\\
Do not explain your reasoning; state only your judgment.\\[0.5em]

\texttt{- Question: \textless question\textgreater}\\[0.25em]
\texttt{- Summary: "\textless document\_summary\textgreater"}\\[0.25em]
\texttt{- Golden answer: \textless gold\_answer\textgreater}\\[0.25em]
\texttt{- User's answer: \textless model\_answer\textgreater}\\[0.25em]
\texttt{- Judgement: }\\
\bottomrule
\end{tabular}
\caption{Prompt used for LLM-based evaluation on narrativeQA.}
\label{tab:prompt_narr}
\end{table*}

\begin{table*}[t]
\centering
\begin{tabular}{p{\linewidth}}
\toprule
\textbf{System Prompt} \\
\midrule
You are the grader who determines the correct answer precisely. \\
\midrule
\textbf{User Prompt} \\
\midrule
Determine whether the user's answer to the following question is correct or not.\\
Choose only one of the two options: \texttt{[Correct]} or \texttt{[Wrong]}.\\
Do not explain your reasoning; state only your judgment.\\[0.5em]

\texttt{- Question: \textless question\textgreater}\\[0.25em]
\texttt{- Literature Context: "\textless chunk\_must\_contain\textgreater"}\\[0.25em]
\texttt{- Golden answer: \textless gold\_answer\textgreater}\\[0.25em]
\texttt{- User's answer: \textless model\_answer\textgreater}\\[0.25em]
\texttt{- Judgement: }\\
\bottomrule
\end{tabular}
\caption{Prompt used for LLM-based evaluation on GutenQA.}
\label{tab:prompt_guten}
\end{table*}

\section{Linking Case Study}
As illustrated in Table~\ref{tab:case}, we present a representative linking case demonstrating how ChronoRAG assembles adjacent passages. 
Bracketed sentences mark the retrieved evidence aligned with the query, while the surrounding context ensures coherence. 
This example shows how linking preserves narrative flow and enables the model to answer correctly (``a dog''), 
reducing ambiguity compared to isolated retrieval.

\begin{table*}[t]
\centering
\begin{tabular}{p{0.9\linewidth}}
\toprule
\textbf{Example Linking Case} \\
\midrule
Correct: [Correct] \\
Question: What kind of animal does Anna have when Gurov first sees her? \\
Golden Answer: [``A dog'', ``Small dog''] \\
Generated Answer: a dog \\
Passage: \\
\quad Dmitri Gurov becomes infatuated with the lady with the dog and tries to get to know her, \\
\quad \textbf{[Gurov is drawn to Anna's innocence and vulnerability, and they spend time together]}, \\
\quad Gurov and Anna share a romantic moment, but Anna is overcome with guilt and shame. \\
\quad \\
\quad Kovrin and the Pesotskys are preparing for the wedding, \\
\quad \textbf{[Kovrin is deeply in love with Tanya, and their relationship is central to the story]}, \\
\quad The black monk appears to Kovrin, filling him with pride and a sense of exalted consequence, \\
\quad and Kovrin is obsessed with his work. \\
\quad \\
\quad Gurov is drawn to Anna's innocence and vulnerability, and they spend time together, \\
\quad \textbf{[Gurov and Anna share a romantic moment, but Anna is overcome with guilt and shame]}, \\
\quad Anna confesses her infidelity to her husband, feeling she has betrayed him. \\
\\
\bottomrule
\end{tabular}
\caption{Representative linking case used in ChronoRAG.}
\label{tab:case}
\end{table*}

\clearpage

\end{document}